\documentclass[conference]{IEEEtran}

\usepackage{amsmath}
\usepackage{tabularx}
\usepackage{booktabs}
\usepackage[utf8]{inputenc}
\usepackage[boxed]{algorithm2e}
\usepackage{float}

\ifCLASSINFOpdf
  \usepackage[pdftex]{graphicx}
\else
\fi

\begin{document}
\title{CloudScan - A configuration-free invoice analysis system using recurrent neural networks}

\author{
  \IEEEauthorblockN{Rasmus Berg Palm}
  \IEEEauthorblockA{DTU Compute\\Technical University of Denmark\\rapal@dtu.dk\\}
  \and
  \IEEEauthorblockN{Ole Winther}
  \IEEEauthorblockA{DTU Compute\\Technical University of Denmark\\olwi@dtu.dk}
  \and
  \IEEEauthorblockN{Florian Laws}
  \IEEEauthorblockA{Tradeshift\\Copenhagen, Denmark\\fla@tradeshift.com}
}

\maketitle

\begin{abstract}
We present CloudScan; an invoice analysis system that requires zero configuration or upfront annotation.

In contrast to previous work, CloudScan does not rely on templates of invoice layout, instead it learns a single global model of invoices that naturally generalizes to unseen invoice layouts.

The model is trained using data automatically extracted from end-user provided feedback. This automatic training data extraction removes the requirement for users to annotate the data precisely.

We describe a recurrent neural network model that can capture long range context and compare it to a baseline logistic regression model corresponding to the current CloudScan production system.

We train and evaluate the system on 8 important fields using a dataset of 326,471 invoices. The recurrent neural network and baseline model achieve 0.891 and 0.887 average F1 scores respectively on seen invoice layouts. For the harder task of unseen invoice layouts, the recurrent neural network model outperforms the baseline with 0.840 average F1 compared to 0.788.

\end{abstract}
%
\IEEEpeerreviewmaketitle

\section{Introduction}

Invoices, orders, credit notes and similar business documents carry the information needed for trade to occur between companies and much of it is on paper or in semi-structured formats such as PDFs \cite{sellen_myth_2003}. In order to manage this information effectively, companies use IT systems to extract and digitize the relevant information contained in these documents. Traditionally this has been achieved using humans that manually extract the relevant information and input it into an IT system. This is a labor intensive and expensive process \cite{klein_results_2004}.

The field of information extraction addresses the challenge of automatically extracting such information and several commercial solutions exists that assist in this. Here we present CloudScan, a commercial solution by Tradeshift, free for small businesses, for extracting structured information from unstructured invoices.

Powerful information extraction techniques exists given that we can observe invoices from the same template beforehand, e.g. rule, keyword or layout based techniques. A template is a distinct invoice layout, typically unique to each sender. A number of systems have been proposed that rely on first classifying the template, e.g. Intellix \cite{schuster_intellix_2013}, ITESOFT \cite{rusinol_field_2013}, smartFIX \cite{dengel_smartfix:_2002} and others \cite{cesarini_analysis_2003,esser_automatic_2012,medvet_probabilistic_2010}. As these systems rely on having seen the template beforehand, they cannot accurately handle documents from unseen templates. Instead they focus on requiring as few examples from a template as possible.

What is harder, and more useful, is a system that can accurately handle invoices from completely unseen templates, with no prior annotation, configuration or setup. This is the goal of CloudScan: to be a simple, configuration and maintenance free invoice analysis system that can convert documents from both previously seen and unseen templates with high levels of accuracy.

CloudScan was built from the ground up with this goal in mind. There is no notion of template in the system. Instead every invoice is processed by the same system built around a single machine learning model. CloudScan does not rely on any system integration or prior knowledge, e.g. databases of orders or customer names, meaning there is no setup required in order to use it.

CloudScan automatically extracts the training data from end-user provided feedback. The end-user provided feedback required is the correct value for each field, rather than the map from words on the page to fields. It is a subtle difference, but this separates the concerns of reviewing and correcting values using a graphical user interface from concerns related to acquiring training data. Automatically extracting the training data this way also results in a very large dataset which allows us to use methods that require such large datasets.

In this paper we describe how CloudScan works, and investigate how well it accomplishes the goal it aims to achieve. We evaluate CloudScan using a large dataset of 326,471 invoices and report competitive results on both seen and unseen templates. We establish two classification baselines using logistic regression and recurrent neural networks, respectively. 

\begin{figure*}[!t]
	\centering \includegraphics[width=1.0\textwidth]{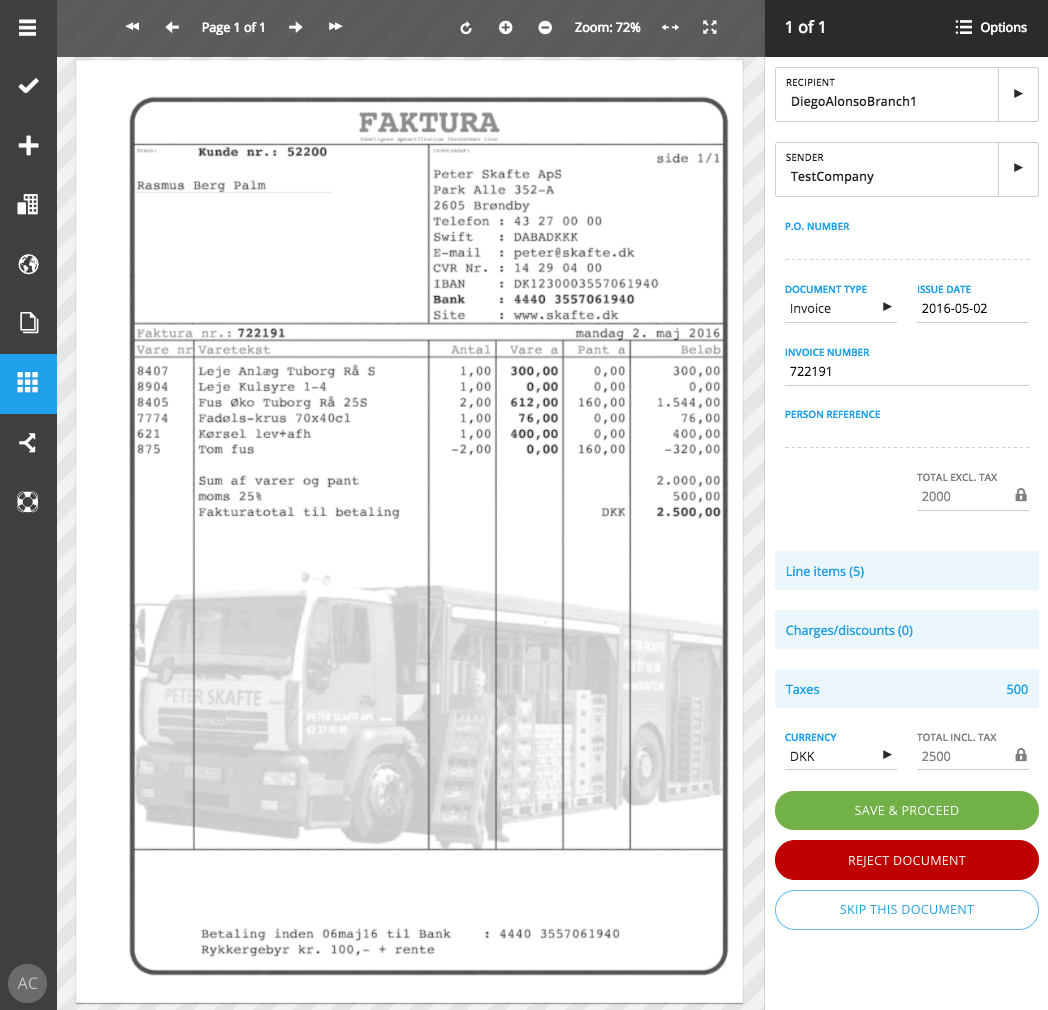}
    \caption{The CloudScan graphical user interface. Results before any correction. Disregard the selected sender and recipient as these are limited to companies connected to the company uploading the invoice. This is an example of a perfect extraction which would give an F1 score of 1.}
	\label{fig:GUI}
\end{figure*}

\section{Related Work}

The most directly related works are Intellix \cite{schuster_intellix_2013} by DocuWare and the work by ITESOFT \cite{rusinol_field_2013}. Both systems require that relevant fields are annotated for a template manually beforehand, which creates a database of templates, fields and automatically extracted keywords and positions for each field. When new documents are received, both systems classify the template automatically using address lookups or machine learning classifiers. Once the template is classified the keywords and positions for each field are used to propose field candidates which are then scored using heuristics such as proximity and uniqueness of the keywords. Having scored the candidates the best one for each field is chosen.

smartFIX \cite{dengel_smartfix:_2002} uses manually configured rules for each template. Cesarini et al. \cite{cesarini_analysis_2003} learns a database of keywords for each template and fall back to a global database of keywords. Esser et al. \cite{esser_automatic_2012} uses a database of absolute positions of fields for each template. Medvet et al. \cite{medvet_probabilistic_2010} uses a database of manually created (field, pattern, parser) triplets for each template, designs a probabilistic model for finding the most similar pattern in a template, and extracts the value with the associated parser.

Unfortunately we cannot compare ourselves directly to the works described as the datasets used are not publicly available and the evaluation methods are substantially different. However, the described systems all rely on having an annotated example from the same template in order to accurately extract information.

To the best of our knowledge CloudScan is the first invoice analysis system that is built for and capable of accurately converting invoices from unseen templates.

The previous works described can be configured to handle arbitrary document classes, not just invoices, as is the case for CloudScan. Additionally, they allow the user to define which set of fields are to be extracted per class or template, whereas CloudScan assumes a single fixed set of fields to be extracted from all invoices.

Our automatic training data extraction is closely related to the idea of distant supervision \cite{mintz_distant_2009} where relations are extracted from unstructured text automatically using heuristics.

The field of Natural Language Processing (NLP) offers a wealth of related work. Named Entity Recognition (NER) is the task of extracting named entities, usually persons or locations, from unstructured text. See Nadeau and Sekine \cite{nadeau_survey_2007} for a survey of NER approaches. Our system can be seen as a NER system in which we have 8 different entities. In recent years, neural architectures have been demonstrated to achieve state-of-the-art performance on NER tasks, e.g.\ Lample et al. \cite{lample_neural_2016}, who combine word and character level RNNs, and Conditional Random Fields (CRFs).

Slot Filling is another related NLP task in which pre-defined slots must be filled from natural text. Our system can be seen as a slot filling task with 8 slots, and the text of a single invoice as input. Neural architectures are also used here, e.g.\ \cite{mesnil_investigation_2013} uses bi-directional RNNs and word embedding to achieve competitive results on the ATIS (Airline Travel Information Systems) benchmark dataset. 

In both NER and Slot Filling tasks, a commonly used approach is to classify individual tokens with the entities or slots of interest, an approach that we adopt in our proposed RNN model.

\section{CloudScan}

\begin{figure*}[!t]
	\centering \includegraphics[width=1.0\textwidth]{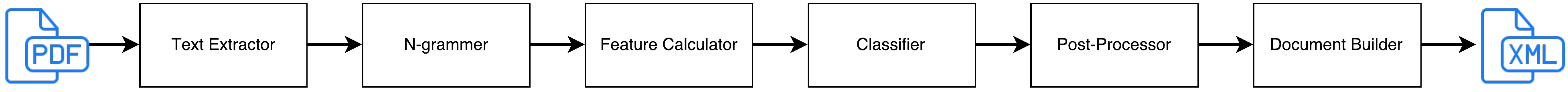}
    \caption{The CloudScan engine.}	
	\label{fig:CloudScan}
\end{figure*}

\subsection{Overview}

CloudScan is a cloud based software as a service invoice analysis system offered by Tradeshift. Users can upload their unstructured PDF invoices and the CloudScan engine converts them into structured XML invoices. The CloudScan engine contains 6 steps. See Figure \ref{fig:CloudScan}.

\begin{enumerate}
\item \textbf{Text Extractor}. Input is a PDF invoice. Extracts words and their positions from the PDF. If the PDF has embedded text, the text is extracted, otherwise a commercial OCR engine is used. The output of this step is a structured representation of words and lines in hOCR format \cite{breuel_hocr_2007}.

\item \textbf{N-grammer}. Creates N-grams of words on the same line. Output is a list of N-grams up to length 4.

\item \textbf{Feature Calculator}. Calculates features for every N-gram. Features fall in three categories: text, numeric and boolean. Examples of text features are the raw text of the N-gram, and the text after replacing all letters with "x", all numbers with "0" and all other characters with ".". Examples of numeric features are the normalized position on the page, the width and height and number of words to the left. Boolean features include whether the N-gram parses as a date or an amount or whether it matches a known country, city or zip code. These parsers and small databases of countries, cities and zip codes are built into the system, and does not require any configuration on the part of the user. The output is a feature vector for every N-gram. For a complete list of features see table \ref{table:features}.

\item \textbf{Classifier}. Classifies each N-gram feature vector into 32 fields of interest, e.g. invoice number, total, date, etc. and one additional field 'undefined'. The undefined field is used for all N-grams that does not have a corresponding field in the output document, e.g. terms and conditions. The output is a vector of 33 probabilities for each N-gram.

\item \textbf{Post Processor}. Decides which N-grams are to be used for the fields in the output document. For all fields, we first filter out N-gram candidates that does not fit the syntax of the field after parsing with the associated parser. E.g. the N-gram "Foo Bar" would not fit the Total field after parsing with the associated parser since no amount could be extracted. The parsers can handle simple OCR errors  and various formats, e.g. "100,0o" would be parsed to "100.00". The parsers are based on regular expressions.

For fields with no semantic connection to other fields, e.g. the invoice number, date, etc. we use the Hungarian algorithm \cite{kuhn_hungarian_1955}. The Hungarian algorithm solves the assignment problem, in which N agents are to be assigned to M tasks, such that each task has exactly one agent assigned and no agent is assigned to more than one task. Given that each assignment has a cost, the Hungarian algorithm finds the assignments that minimizes the total cost. We use 1 minus the probability of an N-gram being a field as the cost.

For the assignment of the Total, Line Total, Tax Total and Tax Percentage we define and minimize a cost function based on the field probabilities and whether the candidate totals adds up.

The output is a mapping from the fields of interest to the chosen N-grams.

\item \textbf{Document Builder}. Builds a Universal Business Language (UBL) \cite{boask_universal_2006} invoice with the fields having the values of the found N-grams. UBL is a XML based invoice file format. Output is a UBL invoice.
\end{enumerate}

\subsection{Extracting training data from end-user provided feedback}
The UBL invoice produced by the engine is presented to the user along with the original PDF invoice in a graphical user interface (GUI). The user can correct any field in the UBL invoice, either by copy and pasting from the PDF, or by directly typing in the correction. See figure \ref{fig:GUI}.

Once the user has corrected any mistakes and accepted the invoice we add the resulting UBL to our data collection. We will extract training data from these validated UBL documents, even though they might deviate from the PDF content due to OCR error, user error or the user intentionally deviating from the PDF content. We discuss these issues later.

The classifier is trained on N-grams and their labels, which are automatically extracted from the validated UBL invoices and the corresponding PDFs. For each field in the validated UBL document we consider all N-grams in the PDF and check whether the text content, after parsing, matches the field value. If it does, we extract it as a single training example of N-gram and label equal to the field. If an N-gram does not match any fields we assign the 'undefined' label. For N-grams that match multiple fields, we assign all matched fields as labels. This ambiguity turns the multi-class problem into a multi-label problem. See Algorithm \ref{alg:data} for details.

\SetAlFnt{\footnotesize}
\SetAlCapSkip{0.5em}
\begin{algorithm}	
    \SetKwInOut{Input}{input}
    \SetKwInOut{Output}{output}
	\SetKwData{nGrams}{nGrams}
    \SetKwData{nGram}{nGram}
    \SetKwData{fields}{fields}
    \SetKwData{field}{field}
    \SetKwData{value}{value}
    \SetKwData{parser}{parser}
    \SetKwData{found}{found}
    \SetKwData{result}{result}
    \SetKwData{parsers}{parsers}
    \SetKwData{maxN}{maxN}
    \SetKwFunction{CreateNgrams}{CreateNgrams}
    \SetKwFunction{GetValue}{GetValue}
    \SetKwFunction{GetParser}{GetParser}
    \SetKwFunction{Parse}{Parse}
    \SetKwFunction{Parse}{Parse}
    \SetKwFunction{Length}{Length}
    \SetKwFunction{Add}{Add}
	\Input{UBL and PDF document}
	\Output{All labeled N-grams}
    \result $\leftarrow \{\}$\;
    \ForEach{\field $\in$ \fields}{
    	\parser $\leftarrow$ \GetParser{\field}\;
        \value $\leftarrow$ \GetValue{UBL, \field}\;
        \maxN $\leftarrow$ \Length{\value} + 2\;
        \nGrams $\leftarrow$ \CreateNgrams{PDF, \maxN}\;
        \ForEach{\nGram $\in$ \nGrams}{
        	\If{\value $=$ \Parse{\nGram, \parser} }{
                \Add{\result, \nGram, \field}\;
            }            
        }        
    }
    \nGrams $\leftarrow$ \CreateNgrams{PDF, 4}\;
    \ForEach{\nGram $\in$ \nGrams}{
    	\If{\nGram $\notin$ \result}{
        	\Add{\result, \nGram, undefined}\;
        }
    }
    \Return{\result}
	\caption{Automatic training data extraction}
    \label{alg:data}
\end{algorithm}

Using automatically extracted pairs like this results in a noisy, but big data set of millions of pairs. Most importantly, however, it introduces no limitations on how users correct potential errors, and requires no training. For instance, we could have required users to select the word matching a field, which would result in much higher quality training data. However in a high volume enterprise setup, this could reduce throughput significantly. Our automatic training data generation decouples the concerns of reviewing and correcting fields from creating training data, allowing the GUI to focus solely on reviewing and correcting fields. The user would need to review the field values and correct potential errors regardless, so as long as we do not limit how the user does it, we are not imposing any additional burdens. In short, the machine learning demands have lower priority than the user experience in this regard.

As long as we get a PDF and a corresponding UBL invoice we can extract training data, and the system should learn and improve for the next invoice.

\section{Experiments}
We perform two experiments meant to test 1) the expected performance on the next invoice, and 2) the harder task of expected performance on the next invoice from an \emph{unseen} template. These are two different measures of generalization performance.

The data set consists of 326,471 pairs of validated UBL invoices and corresponding PDFs from 8911 senders to 1013 receivers obtained from use of CloudScan. We assume each sender corresponds to a distinct template.

For the first experiment we split the invoices into a training, validation and test set randomly, using 70\%, 10\% and 20\% respectively. For the second experiment we split the \textit{senders} into a training, validation and test set randomly, using 70\%, 10\% and 20\% respectively. All the invoices from the senders in a set then comprise the documents of that set. This split ensures that there are no invoices sharing templates between the three sets for the second experiment.

While the system captures 32 fields we only report on eight of them: Invoice number, Issue Date, Currency, Order ID, Total, Line Total, Tax Total and Tax Percent. We only report on these eight fields as they are the ones we have primarily designed the system for. A large part of the remaining fields are related to the sender and receiver of the invoice and used for identifying these. We plan to remove these fields entirely and approach the problem of sender and receiver identification as a document classification problem instead. Preliminary experiments based on a simple bag-of-words model show promising results. The last remaining fields are related to the line items and used for extracting these. Table extraction is a challenging research question in itself, and we are not yet ready to discuss our solution. Also, while not directly comparable, related work \cite{schuster_intellix_2013,rusinol_field_2013,cesarini_analysis_2003} also restricts evaluation to header fields.

Performance is measured by comparing the fields of the generated and validated UBL. Note we are not only measuring the classifier performance, but rather the performance of the entire system. The end-to-end performance is what is interesting to the user after all. Furthermore, this is the strictest possible way to measure performance, as it will penalize errors from any source, e.g. OCR errors and inconsistencies between the validated UBL and the PDF. For instance, the date in the validated UBL might not correspond to the date on the PDF. In this case, even if the date on the PDF is found, it will be counted as an error, as it does not match the date in the validated UBL.

In order to show the upper limit of the system under this measure we include a ceiling analysis where we replace the classifier output with the correct labels directly. This corresponds to using an oracle classifier. We use the MUC-5 definitions of recall, precision and F1, without partial matches \cite{chinchor_muc-5_1993}.

We perform experiments with two classifiers 1) The production baseline system using a logistic regression classifier, and 2) a Recurrent Neural Network (RNN) model. We hypothesize the RNN model can capture context better.

\subsection{Baseline}
The baseline is the current production system, which uses a logistic regression classifier to classify each N-gram individually.

In order to capture some context, we concatenate the feature vectors for the closest N-grams in the top, bottom, left and right directions to the normal feature vectors. So if the feature vector for an N-gram had $M$ entries, after this it would have $5M$ entries.

All $5M$ features are then mapped to a binary vector of size $2^{22}$ using the hashing trick \cite{weinberger_feature_2009}. To be specific, for each feature we concatenate the feature name and value, hash it, take the remainder with respect to the binary vector size and set that index in the binary vector to 1. 

The logistic regression classifier is trained using stochastic gradient descent for 10 epochs after which we see little improvement. This baseline system is derived from the heavily optimized winning solution of a competition Tradeshift held\footnote{https://www.kaggle.com/c/tradeshift-text-classification}.

\subsection{LSTM model}
In order to accurately classify N-grams the context is critical, however when classifying each N-gram in isolation, as in the baseline model, we have to engineer features to capture this context, and deciding how much and which context to capture is not trivial.

A Recurrent Neural Network (RNN) can model the entire invoice and we hypothesize that this ability to take the entire invoice into account in a principled manner will improve the performance significantly. Further, it frees us from having to explicitly engineer features that capture context. As such we only use the original $M$ features, not the $5M$ features of the baseline model. In general terms, a RNN can be described as follows.

\begin{align*}
	h_t &= f(h_{t-1}, x_t) \\
    y_t &= g(h_t)
\end{align*}

Where $h_t$ is the hidden state at step $t$, $f$ is a neural network that maps the previous hidden state $h_{t-1}$, and the input $x_t$ to $h_t$ and $g$ is a neural network that maps the hidden state $h_t$ to the output of the model $y_t$. Several variants have been proposed, most notably the Long Short Term Memory (LSTM) \cite{hochreiter_long_1997} which is good at modeling long term dependencies.

A RNN models a sequence, i.e. $x$ and $y$ are ordered and as such we need to impose an ordering on the invoice. We chose to model the words instead of N-grams, as they fit the RNN sequence model more naturally and we use the standard left-to-right reading order as the ordering. Since the labels can span multiple words we re-label the words using the IOB labeling scheme \cite{ramshaw_text_1995}. The sequence of words "Total Amount: 12 200 USD" would be labeled "O O B-Total I-Total B-Currency".

We hash the text of the word into a binary vector of size $2^{18}$ which is embedded in a trainable 500 dimensional distributed representation using an embedding layer \cite{bengio_neural_2003}. Using hashing instead of a fixed size dictionary is somewhat unorthodox but we did not observe any difference from using a dictionary, and hashing was easier to implement. It is possible we could have gotten better results using more advanced techniques like byte pair encoding \cite{sennrich_neural_2015}.

We normalize the numerical and boolean features to have zero mean and unit variance and form the final feature vector for each word by concatenating the word embedding and the normalized numerical features.

From input to output, the model has: two dense layers with 600 rectified linear units each, a single bidirectional LSTM layer with 400 units, and two more dense layers with 600 rectified linear units each, and a final dense output layer with 65 logistic units (32 classes that can each be 'beginning' or 'inside' plus the 'outside' class).

\begin{figure}[H]
	\label{fig:LSTM}
	\includegraphics[width=0.48\textwidth]{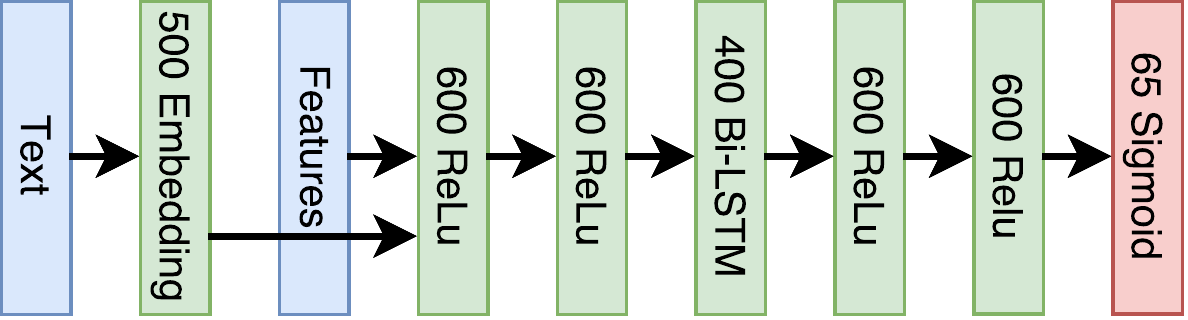}
    \caption{The LSTM model.}
\end{figure}

Following Gal \cite{gal_theoretically_2015}, we apply dropout on the recurrent units and on the word embedding using a dropout fraction of 0.5 for both. Without this dropout the model severely overfits.

The model is trained with the Adam optimizer \cite{kingma_adam:_2014} using minibatches of size 96 until the validation performance has not improved on the validation set for 5 epochs. Model architecture and hyper-parameters were chosen based on the performance on the validation set. For computational reasons we do not train on invoices with more than 1000 words, which constitutes approximately 5\% of the training set, although we do test on them. The LSTM model was implemented in Theano \cite{theano_development_team_theano:_2016} and Lasagne \cite{dieleman_lasagne:_2015}.

After classification we assign each word the IOB label with highest classification probability, and chunk the IOB labeled words back into labeled N-grams. During chunking, words with I labels without matching B labels are ignored. For example, the sequence of IOB labels [B-Currency, O, B-Total, I-Total, O, I-Total, O] would be chunked into [Currency, O, Total, O, O]. The labeled N-grams are used as input for the Post Processor and further processing is identical to the baseline system.

\section{Results}

\begin{table}[!ht]
	\setlength{\tabcolsep}{5pt}
	\renewcommand{\arraystretch}{1.3}
    \caption{Ceiling analysis results. Measured on all documents. Expected performance given an oracle classifier.}
    \label{table:ceiling}
    \centering
	\begin{tabular}{lccc}
    	\toprule
        Field 		& F1 			& Precision		& Recall 	\\
        \midrule
        Number 		& 0.918			& 0.967			& 0.873		\\
        Date		& 0.899			& 1.000			& 0.817		\\
        Currency	& 0.884			& 1.000			& 0.793		\\
        Order ID	& 0.820			& 0.979			& 0.706		\\
        Total 		& 0.966			& 0.981			& 0.952		\\
        Line Total	& 0.976			& 0.991			& 0.961		\\ 
        Tax Total	& 0.959			& 0.961			& 0.957		\\
        Tax Percent	& 0.901			& 0.928			& 0.876		\\
        \midrule
        Micro avg.	& 0.925			& 0.974			& 0.881		\\
        \bottomrule
    \end{tabular}
\end{table}

The results of the ceiling analysis seen in Table \ref{table:ceiling} show that we can achieve very competitive results with CloudScan using an oracle classifier. This validates the overall system design, including the use of automatically generated training data, and leaves us with the challenge of constructing a good classifier.

The attentive reader might wonder why the precision is not 1 exactly for all fields, when using the oracle classifier. For the 'Number' and 'Order ID' fields this is due to the automatic training data generation algorithm disregarding spaces when finding matching N-grams, whereas the comparison during evaluation is strict. For instance the automatic training data generator might generate the N-gram ("16 2054": Invoice Number) from (Invoice Number: "162054") in the validated UBL. When the oracle classifier classifies the N-gram "16 2054" as Invoice Number the produced UBL will be (Invoice Number: "16 2054"). When this is compared to the expected UBL of (Invoice Number: "162054") it is counted as incorrect. This is an annoying artifact of the evaluation method and training data generation. We could disregard spaces when comparing strings during evaluation, but we would risk regarding some actual errors as correct then. For the total fields and the tax percent, the post processor will attempt to calculate missing numbers from found numbers, which might result in errors.

As it stands the recall rate is the limiting factor of the system. The low recall rate can have two explanations: 1) The information is present in the PDF but we cannot read or parse it, e.g. it might be an OCR error or a strange date format, in which case the OCR engine or parsing should be improved, or 2) the information is legitimately not present in the PDF, in which case there is nothing to do, except change the validated UBL to match the PDF.

\begin{table}[!ht]
	\setlength{\tabcolsep}{5pt}
	\renewcommand{\arraystretch}{1.3}
    \caption{Expected performance on next received invoice. Best results in bold.}
    \label{table:ex-1}
    \centering
	\begin{tabular}{lccccccc}
        
    	\toprule
         			& \multicolumn{2}{c}{F1} 		& \multicolumn{2}{c}{Precision}	& \multicolumn{2}{c}{Recall} 	\\
        Field		& Baseline		& LSTM 			& Baseline 		& LSTM 			& Baseline 		& LSTM 			\\ 
        \midrule
        Number 		&\textbf{0.863}	& 0.860			&\textbf{0.883}	& 0.877			&\textbf{0.844}	& 0.843			\\
        Date		& 0.821			&\textbf{0.828}	& 0.876			&\textbf{0.893}	&\textbf{0.773}	& 0.772			\\
        Currency	& 0.869			&\textbf{0.874}	& 0.974			&\textbf{0.992}	&\textbf{0.784}	& 0.781			\\
        Order ID	&\textbf{0.776}	& 0.760			&\textbf{0.936}	& 0.930			&\textbf{0.663}	& 0.642			\\
        Total 		& 0.927			&\textbf{0.932}	& 0.940			&\textbf{0.942}	& 0.915			&\textbf{0.924}	\\
        Line Total	& 0.923			&\textbf{0.936}	& 0.936			&\textbf{0.945}	& 0.911			&\textbf{0.927}	\\ 
        Tax Total	& 0.931			&\textbf{0.939}	& 0.933			&\textbf{0.941}	& 0.929			&\textbf{0.937}	\\
        Tax Percent	& 0.901			&\textbf{0.903}	& 0.927			&\textbf{0.930}	& 0.876			&\textbf{0.878}	\\
        \midrule
        Micro avg.	& 0.887			&\textbf{0.891}	& 0.924			&\textbf{0.930}	& 0.852			&\textbf{0.855}	\\
        \bottomrule
    \end{tabular}
\end{table}

Table \ref{table:ex-1} shows the results of experiment 1 measuring the expected performance on the next received invoice for the baseline and LSTM model. The LSTM model is slightly better than the baseline system with an average F1 of 0.891 compared to 0.887. In general the performance of the models is very similar, and close to the theoretical maximum performance given by the ceiling analysis. This means the classifiers both perform close to optimally for this experiment. The gains that can be had from improving upon the LSTM model further are just 0.034 average F1.

\begin{table}[!ht]
	\setlength{\tabcolsep}{5pt}
	\renewcommand{\arraystretch}{1.3}
    \caption{Expected performance on next invoice from unseen template. Best results in bold.}
    \label{table:ex-2}
    \centering
	\begin{tabular}{lccccccc}
        
    	\toprule
         			& \multicolumn{2}{c}{F1} 		& \multicolumn{2}{c}{Precision}	& \multicolumn{2}{c}{Recall} 	\\
        Field		& Baseline		& LSTM 			& Baseline 		& LSTM 			& Baseline 		& LSTM 			\\ 
        \midrule
        Number 		& 0.711			&\textbf{0.760}	& 0.761			&\textbf{0.789}	& 0.668			&\textbf{0.733}	\\
        Date		& 0.693			&\textbf{0.774}	& 0.759			&\textbf{0.847}	& 0.637			&\textbf{0.712}	\\
        Currency	&\textbf{0.907}	& 0.905			& 0.977			&\textbf{0.983}	&\textbf{0.847}	& 0.838			\\
        Order ID	& 0.433			&\textbf{0.523}	&\textbf{0.822}	& 0.737			& 0.294			&\textbf{0.406}	\\
        Total 		& 0.840			&\textbf{0.896}	& 0.864			&\textbf{0.907} & 0.818			&\textbf{0.884}	\\
        Line Total	& 0.803			&\textbf{0.880}	& 0.826			&\textbf{0.891}	& 0.781			&\textbf{0.869}	\\ 
        Tax Total	& 0.832			&\textbf{0.878}	& 0.835			&\textbf{0.881}	& 0.829			&\textbf{0.874}	\\
        Tax Percent	& 0.812			&\textbf{0.869}	& 0.828			&\textbf{0.887}	& 0.796			&\textbf{0.853}	\\
        \midrule
        Micro avg.	& 0.788			&\textbf{0.840}	& 0.836			&\textbf{0.879}	& 0.746			&\textbf{0.804}	\\
        \bottomrule
    \end{tabular}
\end{table}

More interesting are the results in Table \ref{table:ex-2} which measures the expected performance on the next invoice from an unseen template. This measures the generalization performance of the system across templates which is a much harder task due to the plurality of invoice layouts and reflects the experience a new user will have the first time they use the system. On this harder task the LSTM model clearly outperform the baseline system with an average F1 of 0.840 compared to 0.788. Notably the 0.840 average F1 of the LSTM model is getting close to the 0.891 average F1 of experiment 1, indicating that the LSTM model is largely learning a template invariant model of invoices, i.e. it is picking up on general patterns rather than just memorizing specific templates.

We hypothesized that it is the ability of LSTMs to model context directly that leads to increased performance, although there are several other possibilities given the differences between the two models. For instance, it could simply be that the LSTM model has more parameters, the non-linear feature combinations, or the word embedding.

To test our hypothesis we trained a third model that is identical to the LSTM model, except that the bidirectional LSTM layer was replaced with a feedforward layer with an equivalent number of parameters. We trained the network with and without dropout, with all other hyper parameters kept equal. The best model got an average F1 of 0.702 on the experiment 2 split, which is markedly worse than both the LSTM and baseline model. Given that the only difference between this model and the LSTM model is the lack of recurrent connections we feel fairly confident that our hypothesis is true. The feedforward model is likely worse than the baseline model because it does not have the additional context features of the baseline model.

\begin{table}[!ht]
	\setlength{\tabcolsep}{5pt}
	\renewcommand{\arraystretch}{1.3}
    \caption{Word embedding examples.}
    \label{table:word2vec}
    \centering
	\begin{tabular}{lcc}
    	\toprule
        EUR & GBP & DKK \\
        \$ & USD & DKK \\
        Total & Betrag & TOTAL \\
        Number & No & number \\
        Number: & No & Rechnung-Nr. \\
        London & LONDON & Bremen \\
        Brutto & Ldm & ex.Vat \\
        Phone: & code: & Tel: \\
        \bottomrule
    \end{tabular}
\end{table}

Table \ref{table:word2vec} shows examples of words and the two closest words in the learned word embedding. It shows that the learned embeddings are language agnostic, e.g. the closest word to "Total" is "Betrag" which is German for "Sum" or "Amount". The embedding also captures common abbreviations, capitalization, currency symbols and even semantic similarities such as cities. Learning these similarities versus encoding them by hand is a major advantage as it happens automatically as it is needed. If a new abbreviation, language, currency, etc. is encountered it will automatically be learned.

\section{Discussion}

We have presented our goals for CloudScan and described how it works.
We hypothesized that the ability of a LSTM to model context directly would improve performance.
We carried out experiments to test our hypothesis and evaluated CloudScan's performance on a large realistic dataset.
We validated our hypothesis and showed competitive results of 0.891 average F1 on documents from seen templates, and 0.840 on documents from unseen templates using a single LSTM model. These numbers should be compared to a ceiling of F1=0.925 for an ideal system baseline where an oracle classifier is used.

Unfortunately it is hard to compare to other vendors directly as no large publicly available datasets exists due to the sensitive nature of invoices. We sincerely wish such a dataset existed and believe it would drive the field forward significantly, as seen in other fields, e.g. the large effect ImageNet \cite{russakovsky_imagenet_2015} had on the computer vision field. Unfortunately we are not able to release our own dataset due to privacy restrictions.

A drawback of the LSTM model is that we have to decide upon an ordering of the words, when there is none naturally. We chose the left to right reading order which worked well, but in line with the general theme of CloudScan we would prefer a model which could learn this ordering or did not require one.

CloudScan works only on the word level, meaning it does not take any image features into account, e.g. the lines, logos, background, etc. We could likely improve the performance if we included these image features in the model.

With the improved results from the LSTM model we are getting close to the theoretical maximum given by the ceiling analysis. For unseen templates we can at maximum improve the average F1 by 0.085 by improving the classifier. This corresponds roughly to the 0.075 average F1 that can at maximum be gained from fixing the errors made under the ceiling analysis. An informal review of the errors made by the system under the ceiling analysis indicates the greatest source of errors are OCR errors and discrepancies between the validated UBL and the PDF.

As such, in order to substantially improve CloudScan we believe a two pronged strategy is required: 1) improve the classifier and 2) correct discrepancies between the validated UBL and PDF. Importantly, the second does not delay the turnaround time for the users, can be done at our own pace and only needs to be done for the cases where the automatic training data generation fails. As for the OCR errors we will rely on further advances in OCR technology.

\section*{Acknowledgment}
We would like to thank Ángel Diego Cuñado Alonso and Johannes Ulén for our fruitful discussions, and their great work on CloudScan. This research was supported by the NVIDIA Corporation with the donation of TITAN X GPUs. This work is partly funded by the Innovation Fund Denmark (IFD) under File No. 5016-00101B.

\bibliographystyle{IEEEtran}
\bibliography{zotero}

\begin{table}
	\setlength{\tabcolsep}{5pt}
	\renewcommand{\arraystretch}{1.3}
    \caption{N-gram features.}
    \label{table:features}
    \centering
	\begin{tabularx}{0.50\textwidth}{lX}
    	\toprule
        Name 		& Description 	\\
        \midrule
		RawText & The raw text. \\ 
		RawTextLastWord & The raw text of the last word in the N-gram. \\ 
		TextOfTwoWordsLeft & The raw text of the word two places to the left of the N-gram. \\
        TextPatterns & The raw text, after replacing uppercase characters with X, lowercase with x, numbers with 0, repeating whitespace with single whitespace and the rest with ?. \\ 
		bottomMargin & Vertical coordinate of the bottom margin of the N-gram normalized to the page height. \\ 
        topMargin &  Same as above, but for the top margin. \\
        rightMargin & Horizontal coordinate of the right margin of the N-gram normalized to the page width. \\         
        leftMargin & Same as above but for the left margin. \\
        
		bottomMarginRelative & The vertical distance to the nearest word below this N-gram, normalized to page height. \\ 
        topMarginRelative & The vertical distance to the nearest word above this N-gram, normalized to page height. \\
        rightMarginRelative & The horizontal distance to the nearest word to the right of this N-gram, normalized to page width. \\          
        leftMarginRelative & The horizontal distance to the nearest word to the left of this N-gram, normalized to page width. \\		
        horizontalPosition & The horizontal distance between this N-gram and the word to the left, normalized to the horizontal distance between the word to the left and the word to the right. \\ 
        verticalPosition & Same as above but vertical.\\        
        hasDigits & Whether there are any digits 0-9 in the N-gram. \\ 
		isKnownCity & Whether the N-gram is found in a small database of known cities. \\ 
		isKnownCountry & Same as above, but for countries. \\ 
		isKnownZip & Same as above but for zip codes. \\ 
        leftAlignment & Number of words on the same page which left margin is within 5 pixels of this N-grams left margin. \\
		length & Number of characters in the N-gram. \\ 		
		pageHeight & The height of the page of this N-gram. \\ 
		pageWidth & The width of the page of this N-gram. \\ 
		positionOnLine & Count of words to the left of this N-gram normalized to the count of total words on this line\\ 
		lineSize & The number of words on this line. \\ 
		lineWhiteSpace & The area occupied by whitespace on the line of this N-gram normalized to the total area of the line. \\         
        parsesAsAmount & Whether the N-gram parses as a fractional amount. \\ 
		parsesAsDate & Same as above but for dates. \\ 
		parsesAsNumber & Same as above but for integers. \\
        LineMathFeatures.isFactor & Whether this N-gram, after parsing, can take part in an equation such that it is one of two factors on the same line that when multiplied equals another amount on the same line.  \\
		LineMathFeatures.isProduct & Same as above, except this N-gram is the product of the two factors. \\ 
        \bottomrule
    \end{tabularx}
\end{table}

\end{document}